# Measuring daily-life fear perception change: a computational study in the context of COVID-19[1]

Yuchen Chai
Massachusetts Institute of Technology
MA, USA
ycchai@mit.edu

Juan Palacios
Massachusetts Institute of Technology
MA, USA
jpalacio@mit.edu

Jianghao Wang
Chinese Academy of Science
Beijing, China
wangjh@lreis.ac.cn

Yichun Fan
Massachusetts Institute of Technology
MA, USA
ycfan@mit.edu

Siqi Zheng
Massachusetts Institute of Technology
MA, USA
sqzheng@mit.edu

## ABSTRACT

COVID-19, as a global health crisis, has triggered the fear emotion with unprecedented intensity. Besides the fear of getting infected, the outbreak of COVID-19 also created significant disruptions in people's daily life and thus evoked intensive psychological responses indirect to COVID-19 infections. Here, we construct an expressed fear database using 16 million social media posts generated by 536 thousand users between January 1st, 2019 and August 31st, 2020 in China. We employ deep learning techniques to detect the fear emotion within each post and apply topic models to extract the central fear topics. Based on this database, we find that sleep disorders ("nightmare" and "insomnia") take up the largest share of fear-labeled posts in the pre-pandemic period (January 2019-December 2019), and significantly increase during the COVID-19. We identify health and work-related concerns are the two major sources of fear induced by the COVID-19. We also detect gender differences, with females generating more posts containing the daily-life fear sources during the COVID-19 period. This research adopts a data-driven approach to trace back public emotion, which can be used to complement traditional surveys to achieve real-time emotion monitoring to discern societal concerns and support policy decision-making.

## 1. INTRODUCTION

Fear is one of the six basic emotions [1], which is commonly considered to be a brief episode of response to a given threat, either physical or psychological [2,3]. Fear is not merely generated by the direct exposure to a threat to oneself [4]. It can also be transmitted indirectly through social transmission [5]. The perception of fear influences the decision-making process [4] and ultimately translates into behavioral change to help individuals avoid or confront the threat [6,7,8]. However, besides its benefits, fear could also lead to chaos in society. For instance, panic buying is a typical response to the uncertainty of crises, which depletes public resources rapidly and unnecessarily [9]. In other cases, the emotion of fear evoked by the social and political environment could lead to violence and protests [10].

Against this background, it is crucial for policy-makers to understand the causes and development of fear to solve the problems and calm down public anxiety [2]. With a good understanding of the sources and evolution of fear emotion, this seemingly negative emotion could even serve as a valuable tool for public agencies to promote socially desirable actions, such as conservation behaviors to mitigate climate change [11] and social distancing behaviors during epidemics [12].

Previous studies have found that disasters and crises could induce fear emotion [13]. COVID-19, as a global health crisis, has triggered fear vastly with unprecedented intensity [14,15]. Such fear is not merely driven by virus infection. As the pandemic is accompanied by the implementation of non-pharmaceutical policy interventions, the outbreak of COVID-19 has created enormous impacts on people's daily life and evokes intensive psychological responses not directly related to COVID-19 infections [16,17].

Researchers and policy-makers mainly rely on surveys to measure fear perception [18]. However, surveys have their limitations, such as limited scalability, potential sample bias, high cost, and significant time delays [19]. These drawbacks are especially prominent in the context of COVID-19 when the public sentiment evolved rapidly, and timely interventions are critical for lives. When coupled with machine learning techniques, social media platforms could serve as a valuable tool, which enables the monitoring of public emotions with high temporal and spatial granularity. For example, using social media posts, Dodds et al. [20] explored the temporal pattern of emotions for 63 million users non-invasively; Mitchell et al. [21] estimated geographical happiness distribution using the geotagged Twitter. A recent study also shows the high correlation between social media expressed emotion measurement and traditional surveys [22], supporting the validity of such NLP methods to measure emotion.

In this paper, we study how COVID-19 triggered fear for different aspects of people's daily life (the contents that people posted not directly mentioning the virus-related words). This is achieved by using the Bidirectional Encoder Representations from Transformers (BERT) model to detect fear contents and the BERTopic to extract fear-related topics on a self-constructed social media dataset.

---

## 2. METHODS
### 2.1 Data Collection and Preprocessing
We collect the social media data from Sina-Weibo's (the largest microblogging social media platform in China) application programming interface (API). The data contains 16 million original posts from a cohort of 536 thousand active users between January 1st, 2019, and August 31st, 2020.

Besides the raw content, we collect the exact posting time, number of likes, and re-posts for each post. To ensure data quality, we follow several rules when collecting data and constructing the research database: 1) We only collect posts from those users who registered before January 1st, 2019; 2) We exclude the posts generated by institutional accounts (e.g., companies and organizations) from our sample; 3) We drop users with post numbers within the top 10% to reduce the influence from extreme posters; 4) We randomly select and scrutinize 50 thousand posts to identify advertisements with a fixed format (For instance: I am the 3545th to celebrate the shopping festival, please join us!). We then apply regular expressions to remove advertisements in these formats for all posts; 5) We apply a series of functions to remove URLs, emojis, special characters, hash symbols from the posts to reduce the impacts of irrelevant information.

We retrieve the publicly accessible personal information from the profile page of each individual in our sample, including the birth date, gender, number of fans, number of followers, and the registration location. Supplementary Table A.1 shows the summary statistics of our users. All people provide gender information, with 65.31% users reported to be a woman. 63.0% users provide birth date, with average age of 29.01 (SD = 5.85, Min = 10, Max = 80). Supplementary Figures A.2 and A.3 show the comparison of location and age distribution between Weibo users and Chinese 2010 census data.

### 2.2 Expressed Fear Emotion Classification Using Natural Language Processing
Natural Language Processing (NLP) is a computational method that translates unstructured large-scale text data into structured measures [23]. Sentiment analysis, a sub-area of NLP, is purposefully designed to evaluate the emotional status embedded in the text [24]. An increasing number of studies attempt to detect the change of perceptions or attitudes on social media either towards general or specific topics based on the measures generated from these methods [25].

In this study, we use BERT, a text classification model developed by Google [26], to classify each post into six categories of emotions (i.e., Anger, Fear, Happiness, Sadness, Surprise, and Others). Specifically, we finetune a pre-trained BERT model provided by [27] using our data and then impute the likelihood of expressing emotion in each post for each of the six emotions. The posts are tagged with the emotion of the highest possibility. Supplementary Section B.1 describes the construction of the training dataset according to Lyu et al. [28]; Supplementary Figure B.2 displays the performance of our trained model in classifying fear posts; Supplementary Figure B.3 shows the temporal trend of the six emotions, which depicts that the major emotional response during COVID-19 is the fear emotion.

To better understand how the fear in topics not directly related to COVID-19 developed, we construct a dictionary of COVID-19 related words (See Supplementary Table B.4). The post that contains any word in the list will be treated as COVID-19 related posts. Supplementary Figure B.5 shows how fear posts classified as COVID-19 and non-COVID-19 related evolved on a daily basis.

### 2.3 Topic Modeling
To understand why people express fear emotion in social media, we implement the topic model to discover the abstract topics within the dataset. Such a method is widely used by researchers to understand the public attentions and opinions [29,30]. BERTopic is a state-of-the-art machine learning method that leverages BERT embeddings, uniform manifold approximation and projection (UMAP) dimensionality reduction, hierarchical density-based spatial clustering of applications with noise (HDBSCAN), and class-based term frequency-inverse document frequency (c-TF-IDF) [31] to identify interpretable topics. Using a pre-trained multi-lingual sentence embedding model to encode the text, we apply BERTopic on non-COVID-19 fear posts to identify the fear sources in people's daily life. We apply the model on COVID-19 posts as well to support the analysis. To decide the best topic size, we impute the coherence score by varying the number of clusters. As shown in Supplementary Figure C.1, topic sizes of 60 and 30 are chosen for the two groups respectively. We re-run the algorithm and display the most informative words of each topic (using c-TF-IDF) in Supplementary Table C.2.1 and C.2.2. For sample posts in topic clusters, please refer to Supplementary Table C.3.

## 3. RESULTS
### 3.1 Emotion classification
We find that fear emotion is relatively stable across 2019. There are 2.45% posts on average classified as fear posts (i.e., posts dominated by fear emotion) for each day, while the share reaches a peak of 9.1% on January 23rd (the date that epi-center Wuhan city announced lockdown). The share of fear posts drops afterward and remains 2.64% of total posts after April 8th, 2020, slightly higher than the 2019 baseline.

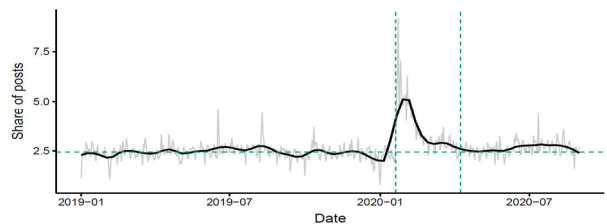

Figure 1: Line graph shows the daily trend of the share of fear posts among all posts. Light grey and the dark black line show the original and smoothed time series respectively. To better locate the peak COVID-19 period, we draw two vertical dashed lines in the plot showing the start of COVID-19 (left, January 20th) and the re-open date of Wuhan city (right, April 8th). The horizontal dashed line depicts the average share of posts during the year 2019.

### 3.2 Evolution of non-COVID-19 related fear topics
BERTopic automatically splits the data into meaningful clusters. There are 60 fear topics unrelated to COVID-19, lying into six large categories (See Supplementary Table C.2.2 and Figure D.1). Health-related fear topics take up the largest share among all the fear posts, followed by relationship, weather and catastrophe, transportation and work/ education. To identify the fear alterations, we conduct t-tests to compare the fear share by topics during and after the peak COVID-19 pandemic with the same period in 2019. Specifically, we define the pandemic periods in China as follows: (1) COVID-19 peak period started from January 20th, 2020 and ended on April 8th, 2020 (i.e., the date when Wuhan city, the epi-center of COVID-19 pandemic in China, re-opened); (2) post-COVID-19 period started from April 9th and ended at August 31st for 2020. Health and work-related topics had the largest change during the COVID-19.

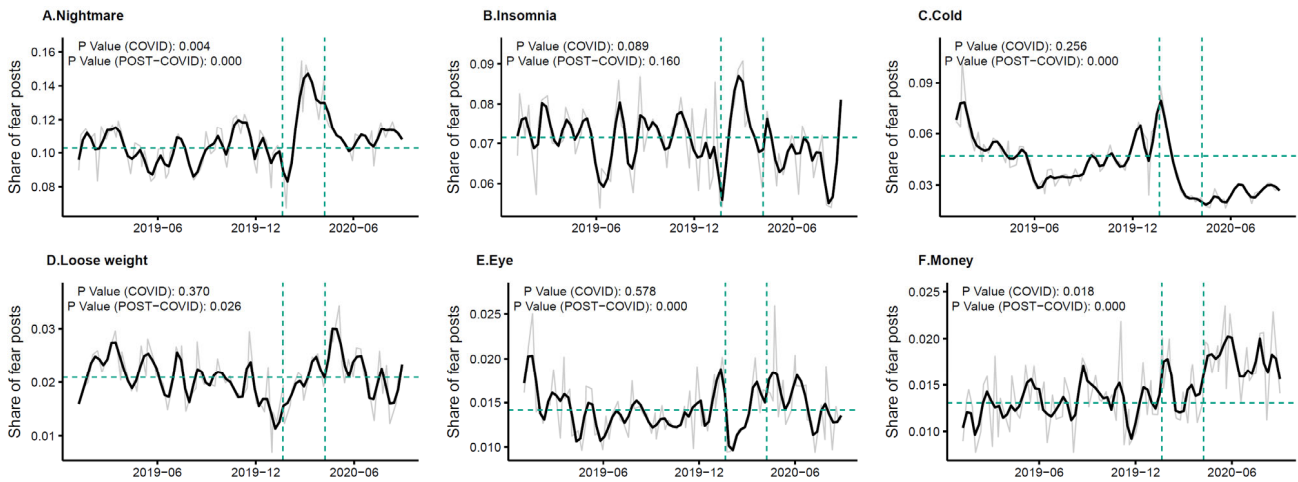

**Figure 2:** Line graphs show the number of posts for nine non-COVID-19 related topics by week. The name of each subplot is the most informative word for each topic. The dark solid lines in each subplot display the smoothed number of posts per day. P-value (COVID) and P-value (post-COVID) indicate the t-test results testing the differences of trend between 2020 peak COVID-19/ post-COVID-19 periods with the same period in 2019.

For health-related topics, we find that topics regarding sleep (i.e., nightmare and insomnia) have the largest share in fear posts during our sample period of two years. On average, 10% and 7% of fear posts are related to nightmares and insomnia, respectively. As shown in Figure 2A, during the COVID-19 peak period, fear posts with contents of "nightmare" significantly increased, reaching a share of 16% of all fear posts. Though this share dropped after the COVID-19 peak period, it remains significantly higher than the same period in 2019 until the end of August, indicating a long-lasting impact. Since "nightmare" could be expressed not only as having an unpleasant dream but also as a way to describe a disastrous event, we further explore the posting time within a day to check whether the fear posts are likely to be sleep-related. We assume that if the "nightmare" is used to describe the awful dream, people are more likely to post in the morning right after having a bad sleep. The results in Supplementary Figure D.2 indeed show that posts about "nightmare" are concentrated in the early morning, and the posting times within a day are similar in 2019 and 2020, indicating that there is no significant change in word usage. "Insomnia", i.e., unable to sleep, displays a similar spike during the COVID-19 peak period (Figure 2B), suggesting that people were more likely to have difficulty falling asleep. However, the share of "insomnia" posts soon recovered to pre-pandemic status after April. Besides sleep disorders, among health topics, we also notice a significant drop in posts mentioning "cold and fever" (Figure 2C), and a significant increase in posts mentioning "lose weight" (Figure 2D) and "eye" (Figure 2E).

Besides health, work is one of the key areas for which the COVID-19 pandemic created significant impacts. Many researchers have identified the economic impacts of COVID-19 [14,32]. The lockdown policy could curb the infections but at the same time prevent people from going to work. The share of posts mentioning "money" increased significantly since the beginning of the COVID-19, suggesting the rises in financial concerns. After checking the content of posts, we find that people are paying more attention to the importance of having money imposed by the pandemic.

### 3.3 Gender difference

Researchers have identified significant gender differences during the COVID-19 period in aspects such as risk perception, time use, and compliance to social distancing policies [33,34]. Here, we explore the difference between genders regarding fear perceptions and topics. Females in general have a higher tendency to express fear. On average, a female user generates 2.27 fear posts during our research period, while a male user only generates 1.89 fear posts. For each topic, we apply four t-tests to detect the gender differences in the COVID-19 peak period and the post-COVID-19 period between 2019 and 2020 (see Supplementary Table D.4)

Regarding the fear related to "nightmare", we find that both genders increase posting during the COVID-19 period, with females having a larger and more significant extent (coefficient = 0.251, P-value = 0.005) comparing to males (coefficient = 0.103, P-value = 0.193). After the COVID-19 peak period, both genders remain to have a significantly higher frequency of nightmare-related fear posts (with coefficients of 0.270 and 0.197 for females and males respectively). The insomnia topic also shows a similar pattern that the female had a significant increase in posting during the COVID-19 period (coefficient = 0.1, P-value = 0.057). The results from the two sleep-related topics suggest that females are more likely to have sleep disorders during the COVID-19. And such impact lasts for months.

We also detect the differential changes by gender in the "cold and fever" topic. Cold and fever are prevalent in winter seasons, as shown by the peaks at the beginning of 2019 and 2020. However, unexpectedly, the post number drops quickly during the COVID-19 period. Females reduce the posts related to the "cold and fever" topic for non-COVID-19 related posts more than males. The reason for such difference is that females have a higher tendency to associate cold and fever symptoms to COVID-19, which again, reflects potentially higher mental stress of females during the pandemic.

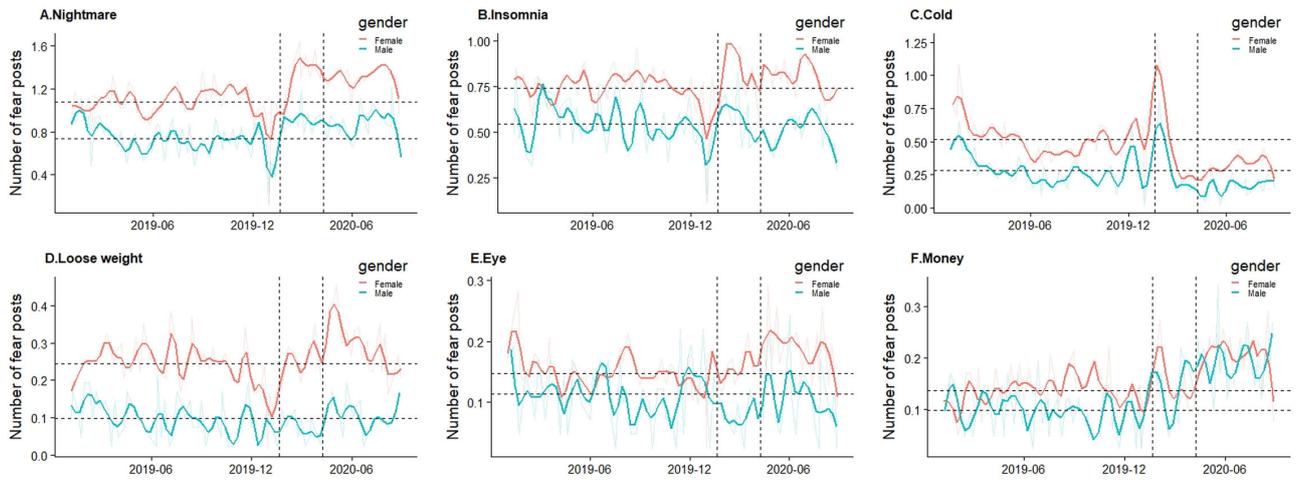

**Figure 3: Line graphs show the average number of fear posts generated by every 1,000 users in each gender by week (Female: solid line above, Male: solid line below). Two horizontal dashed lines depict the baselines (the mean values of 2019) by gender. Two vertical dashed lines show the start date of COVID-19 (January 20th) and the re-open date of Wuhan (April 8th).**

Another pattern we find is related to losing weight. Males post less during the COVID-19 period while females increase their posting after the COVID-19 peak period in this topic. This suggests that people in our sample were less concerned about body shape during the peak pandemic period yet soon start to pay more attention to it once they need to resume work and social activities. The increasing concerns for weight loss could also indicate a reduction in physical activity, as found in previous studies [34].

Regarding monetary topics, both males and females increase their posting behavior during the COVID-19 period, with males having a larger extent (coefficient = 0.042, P-value = 0.064) comparing to females (coefficient = 0.034, P-value = 0.051). Such a concern becomes more significant after the COVID-19 peak period (Male coefficient = 0.090, P-value = 0.000; Female coefficient = 0.062, P-value = 0.000). The work-related topic result shows that, in opposite to health-related topics, males pay more attention to the economic side, indicating a different type of stress. The result could serve as a potential explanation of why men are having a higher suicide rate during the COVID-19 period [35].

## 4. DISCUSSION

In conclusion, our study shows that the COVID-19 has altered people's fear perception towards daily life topics unrelated to virus infection, and the perception change can last for months after the peak pandemic period. We find that the daily-life fear topics in the COVID-19 period which has significant change can be best classified into three clusters: (1) symptoms of fear (such as "nightmare", "insomnia"), (2) fear related to other health problems (such as "lose weight", "eye"), (3) fear about socio-economic consequences (such as "money").

Our results have important implications. First, the significant increases in fear towards these topics indicate an increase in the mental distress and anxiety caused by the COVID-19. Our result shows that fear posts related to "nightmare", the largest non-COVID-19 related fear source, take up a significantly higher proportion of fear posts even months after the peak pandemic. Deteriorated sleep quality brought by mental distress during the COVID-19 could contribute to latent risks for the population's physical and psychological health, which should receive added attention. Second, our results suggest that COVID-19 and related policies induced health and financial concerns. Staying at home was accompanied by a reduction in physical activities and an increase in screen time, thus inducing more fear posts for weight and eye problems. The increased attention to "money" indicates that people were also faced with higher economic burdens during the pandemic. These results reveal the importance of paying attention to the broader social consequences of the COVID-19 on people's daily life, instead of solely focusing on the COVID-19 related posts when analyzing the fear response. Finally, our findings indicate that females are more affected by the COVID-19 in general while males are more concerned with work-related issues points out the importance to explore further the reasons that underlie the sub-group differences in fear responses. Such investigations can assist the designs of tailored policies for the vulnerable population.

Our work leverages the large-scale social media data coupled with computational methods to track the emotional response on a larger scale and with higher temporal granularity than the traditional surveys. Although we conduct this research in a tracing back mode, it is possible to use such a method to achieve real-time emotion monitoring, thus serving as a helpful tool to discern societal concerns and aid for policy decision-making. Our method also has several limitations. First, users of social media platforms might not be able to represent the whole population. Research has found that social media users are younger and are more concentrated in big cities [36] which we also observe in our sample. Second, we use the expressed fear within posts to proxy the fear emotion. Whether the expressed emotion could accurately represent the inner emotional state is still a nascent research area and thus without a clear conclusion. Third, even if the expressed fear can represent the actual feeling of users, we only observe changes in the number of posts with fear as the dominant emotion. Our algorithm does not directly measure the fear intensity of each post at the current stage. Fourth, comparing to a delicately designed survey, using the data-driven method to automatically extract information from unstructured social media posts has unavoidable measurement errors, since the neural network can only capture the general knowledge from training samples and neglects the varying outliers. We hope that our work can motivate more future studies to explore the value of computational methods to understand human emotions and behaviors.

# A WEIBO USER STATISTICS

## A.1 Weibo user statistical information

Table A.1: Collected Weibo User Statistical Information

| Variable | Observations | Mean | St. Dev | Min | Max |
| --- | --- | --- | --- | --- | --- |
| Age | 337,579 | 29.07 | 5.85 | 10 | 81 |
| Female (1 = Yes) | 536,153 | 0.653 | 0.48 | 0 | 1 |
| Number of fans | 536,153 | 4403 | 149288 | 0 | 56407118 |
| Number of followers | 536,153 | 402 | 477 | 0 | 20000 |

## A.2 Location distribution comparison between Weibo user and 2010 census

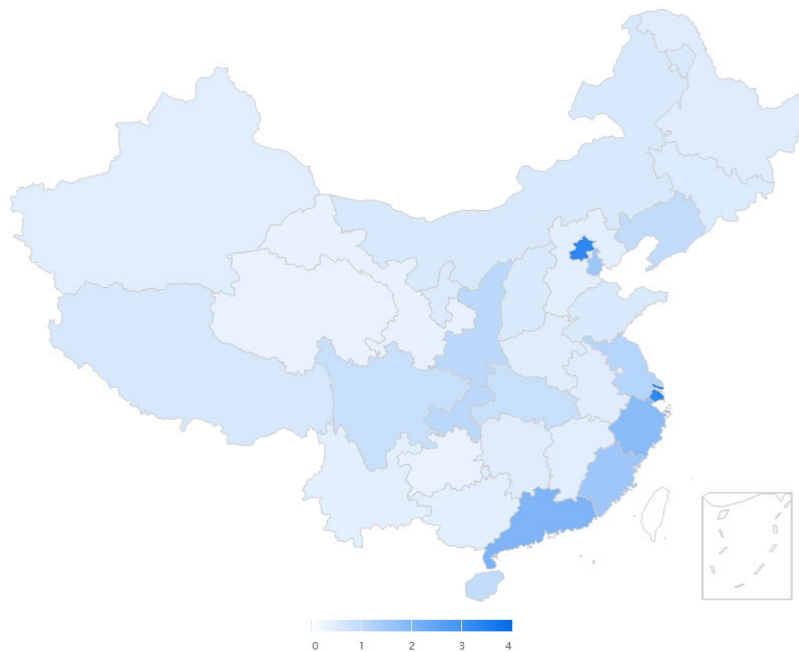

Figure A.2: Provincial map shows the oversampling rate for provinces in China. A more saturated color represents the larger ratio of relative proportion for each province of Weibo users comparing to that of the 2010 census data. Beijing (oversampling rate = 6.53) and Shanghai (oversampling rate = 3.25) are the two most oversampled provinces, followed by provinces in the east of China. To better display the relative relationships between provinces, we truncate the value at 4 to color the map.

## A.3 Age distribution comparison between Weibo user and 2010 census

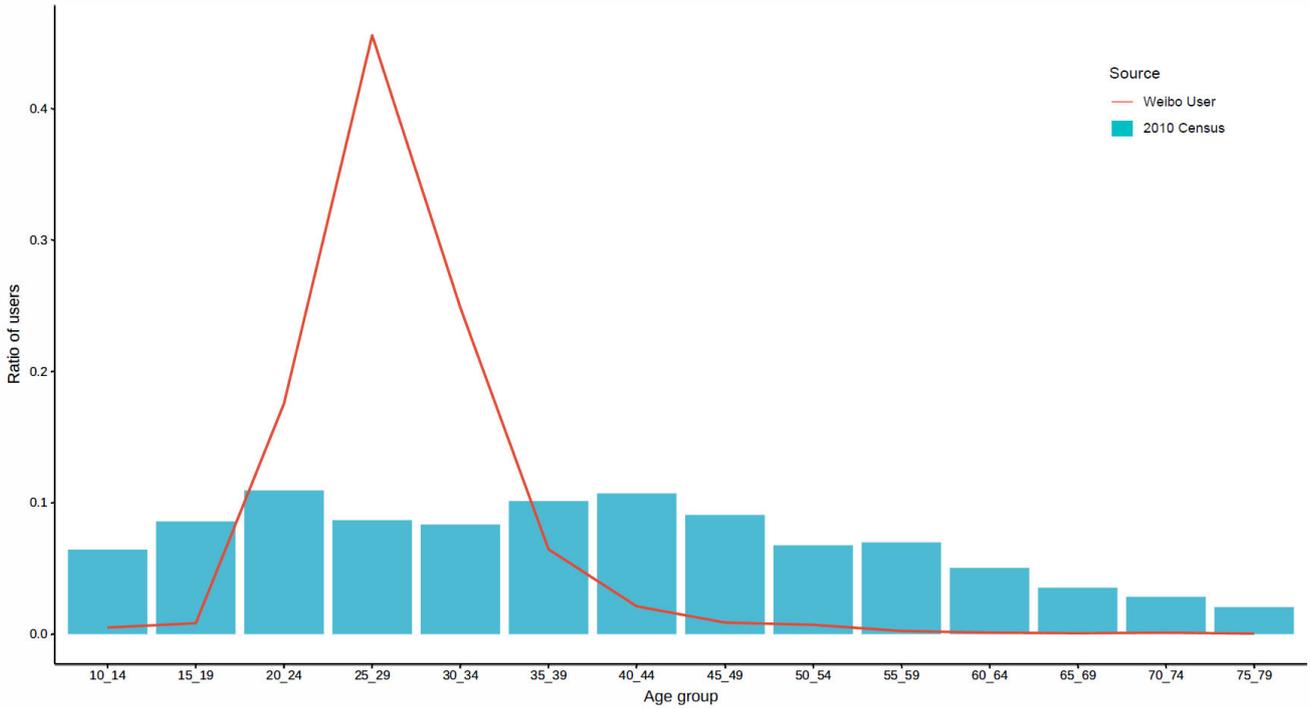

Figure A.3: The age distribution figure compares the Weibo users with 2010 census data. Blue bars in the plot display the age distribution of census while the red line represents Weibo users. Weibo users are more concentrated at a younger age range between 20-40, indicating the disproportionate distribution.

# B MEASURING EXPRESSED EMOTIONS IN WEIBO POSTS

## B.1 Dataset construction

According to Lyu et al. [28], we constructed a multi-class emotion dataset to train the BERT model that consists of the following three parts, including Natural Language Processing and Chinese Computing (NLPCC) emotion analysis dataset, the Evaluation of Weibo Emotion Classification Technology of Social Media Processing 2020 (SMP2020-EWECT), and our dataset. Though we focus on the fear emotion, we still reserve five other emotions, i.e. anger, surprise, happiness, sadness, and others. We construct our labeling dataset by combining two publicly available datasets labeling tweets with emotion tags, and one self-constructed dataset:

### B.1.1 NLPCC emotion analysis dataset

The hosts of NLPCC2014 released thousands of posts (composed of multiple sentences) collected from Weibo. For each sentence within a post, there is a manually labeled tag indicating the expressed emotion, i.e., anger, disgust, fear, happiness, like, sadness, surprise, and others. To be consistent with the SMP2020-EFFECT dataset, we drop the disgust tag and convert the like tag into happiness. Finally, we get 45 thousand labeled sentences.

### B.1.2 SMP2020-EWECT dataset

The SMP2020-EWECT is aimed to detect the emotion within each Weibo post during the COVID-19 period. With each post labeled as one of the tags among anger, fear, happiness, sadness, surprise, and others, it contains two topics which are usual topics and COVID-19 topics. For each topic, the host provides training, evaluation and testing datasets. We leave the testing datasets of usual topics and COVID-19 topics for the model evaluation and combine the rest datasets as training. Finally, we get 40 thousand labeled sentences for training and 5 thousand, 3 thousand for each topic to test.

### B.1.3 Self-constructed dataset

Due to the unbalanced distribution of emotions in the post and lack of training samples, we decide to build our dataset as well. We randomly selected 10 thousand unique posts from the preprocessed post dataset and assign 5 thousand posts to each of the two hired RAs without duplication to manually label them. RAs were told to label any post they could discern fear, anger, sadness, or surprise. To validate the quality of labeling, a researcher went over every post being selected by the RAs and removed any disagreed posts. Finally, there are 1,486 fear posts, 705 anger posts, 652 sadness posts, and 416 surprise posts that passed the quality check.

Three sources of datasets are combined to be one. Since the number of posts in each category of emotion is different and the fear posts have a minimum number of 3,719, we constructed a balanced dataset by randomly selecting 3,719 posts from each category.

We used 22.3 thousand posts in total for training, with 20% are for validation and additional 8 thousand posts in two topics for testing. It is worth noting that the 8 thousand testing posts are provided by SMP2020-EWECT with two sub-topics, i.e., topics non-related to COVID-19 and topics related to COVID-19.

### B.2 Trained model performance evaluation

Results show that the model achieves 74.43% accuracy on the validation dataset. On testing datasets, the overall accuracy is 75.84% (usual topics) and 74.00% (COVID-19 topics). The fear detection achieves 84% and 74% respectively.

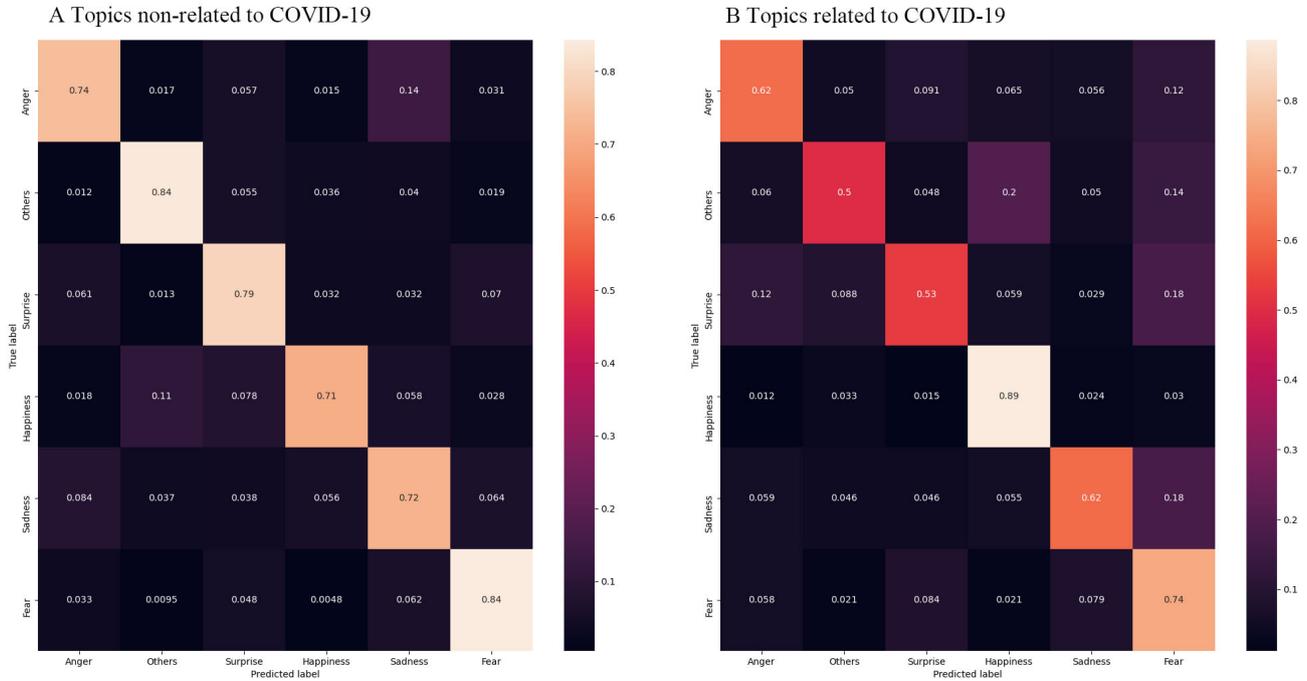

**Figure B.2: Confusion matrixes display the performance of the deep learning model of detecting each of the six emotions considered in the study. Panel A (left) shows the performance in topics that do not relate to COVID-19, and Panel B (right) displays the performance in topics that relate to the COVID-19 pandemic.**

### B.3 Temporal trend of emotions

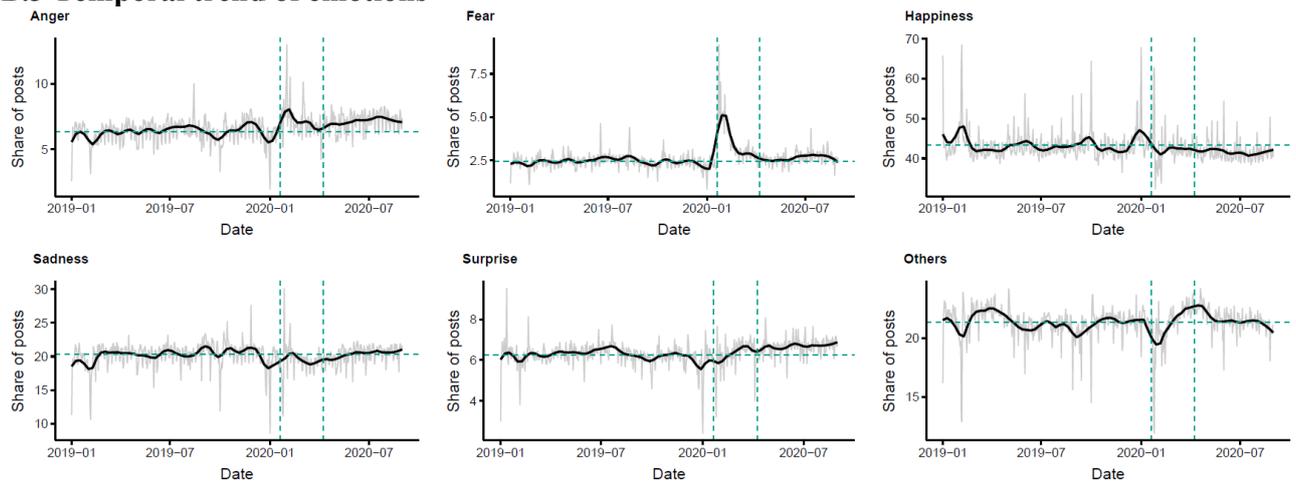

**Figure B.3: Line graphs show the trend of identified six emotions of our data. On average, 42% of the posts per day contain happiness and 20% of the posts contain sadness. Fear is the smallest emotion category with 2.5% posts per day while having a significant increase during the COVID-19 period.**

## B.4 COVID-19 related words

Table B.4: COVID-19 related words

| Index | Word | Translation | Index | Word | Translation |
|---|---|---|---|---|---|
| 1 | 疫情 | pandemic | 13 | 隔离 | isolation |
| 2 | 防控 | control | 14 | 确诊 | confirmed case |
| 3 | 新型 | new type | 15 | 疑似 | suspected |
| 4 | 冠状 | coronal | 16 | 病例 | case |
| 5 | 病毒 | virus | 17 | 告急 | in danger |
| 6 | 爆发 | break out | 18 | 疫苗 | vaccine |
| 7 | 肺炎 | pneumonia | 19 | 封城 | lockdown |
| 8 | 医护 | medical care | 20 | 解封 | lift restriction |
| 9 | 抗疫 | anti-virus | 21 | 湖北 | hubei |
| 10 | 口罩 | mask | 22 | 武汉 | wuhan |
| 11 | 感染 | infection | 23 | 核酸 | nucleic acid |
| 12 | N95 | N95 | 24 | | |

## B.5 Temporal trend of COVID-19 and non-COVID-19 related fear posts

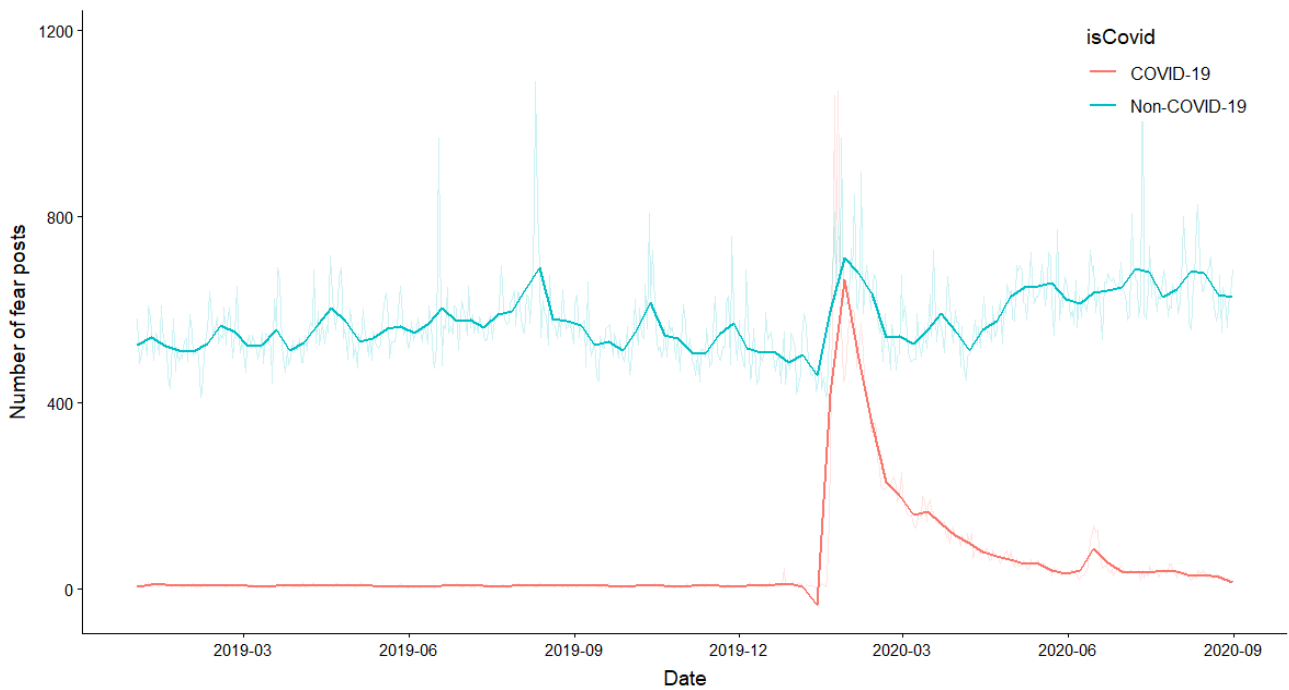

**Figure B.5:** Temporal trend displays the number of COVID-19 and non-COVID-19 posts per day. The line on the bottom represents COVID-19 related post which started to increase on January 20th, reached the peak on January 23rd and gradually decreased after that. The upper line depicts the trend of non-COVID-19 posts, fluctuating around 573 fear posts per day.

# C TOPIC MODELING OF FEAR POSTS
## C.1 Coherence score of topic cluster by varying topic size

A Topics related to COVID-19

B Topics non-related to COVID-19

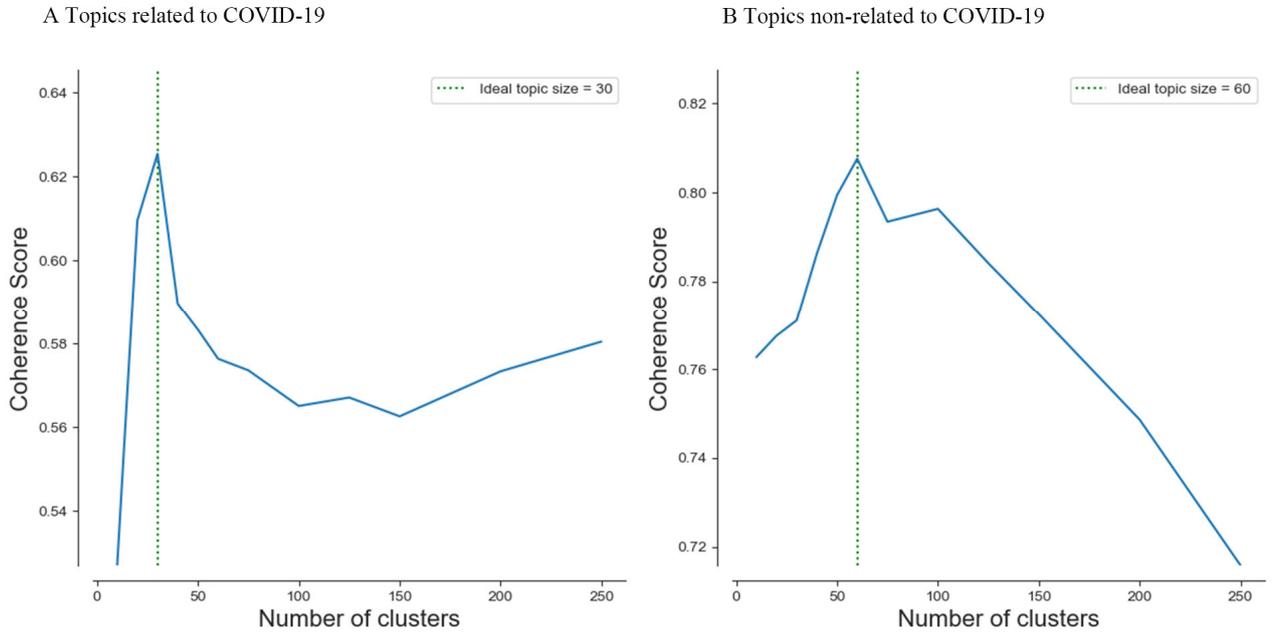

Figure C.1: Coherence score displays the ideal number of clusters for two groups of texts. Panel A (left) shows the highest coherence score for topic clustering on COVID-19 related posts could be achieved when taking 30 as the number of clusters. Panel B (right) shows that setting topic size as 60 would be ideal for non-COVID-19 related posts.

## C.2 Most informative words of topics

Table C.2.1: COVID-19 related topics

| Topic | Size | Word 1 | Word 2 | Word 3 |
|---|---|---|---|---|
| 1 | 4204 | epidemic | the end | past |
| 2 | 1701 | isolation | 14 | nucleic acid |
| 3 | 920 | masks | go out | naked running |
| 4 | 891 | masks | cannot buy | pharmacy |
| 5 | 875 | baby | fever | hospital |
| 6 | 753 | virus | terrible | infection |
| 7 | 677 | masks | allergy | ear |
| 8 | 532 | dream | dreamed of | in the dream |
| 9 | 515 | wuhan city | four people | hubei province |
| 10 | 508 | wuhan city | lockdown | hope |
| 11 | 457 | materials | emergency | request for help |
| 12 | 354 | united states | diagnose | covid |
| 13 | 314 | hope | past | epidemic |
| 14 | 311 | wuhan city | rainstorm | the weather |
| 15 | 303 | pneumonia | new type | covid |
| 16 | 250 | 2020 | kobe bryant | spring festival |
| 17 | 245 | epidemic | terrible | anxiety |
| 18 | 243 | subway | masks | stop |
| 19 | 216 | diagnose | grand total | case |
| 20 | 210 | cannot sleep | insomnia | go to bed |
| 21 | 207 | vaccine | rabies | nine-valent |
| 22 | 203 | aircraft | flight | airport |
| 23 | 183 | wuhan city | virus | infection |
| 24 | 182 | holiday | go to work | jobs |
| 25 | 182 | past | epidemic | faster |

| Topic | Size | Word 1 | Word 2 | Word 3 |
|---|---|---|---|---|
| 26 | 181 | lockdown | reopen | don't understand |
| 27 | 180 | vaccine | kitty | a |
| 28 | 163 | pneumonia | wuhan city | covid |
| 29 | 156 | china | country | epidemic |
| 30 | 152 | n95 | masks | medical |

Table C.2.2: Non COVID-19 related topics

| Topic | Size | Word 1 | Word 2 | Word 3 |
|---|---|---|---|---|
| 1 | 17468 | nightmare | dream | bad dream |
| 2 | 11410 | insomnia | cannot sleep | go to bed |
| 3 | 6705 | cold | fever | cough |
| 4 | 6366 | drive a car | driver | high speed |
| 5 | 5271 | afraid | lonely | terrible |
| 6 | 4935 | dizziness | height | head |
| 7 | 4732 | rain | rainstorm | heavy rain |
| 8 | 4320 | the film | terror | horror film |
| 9 | 4292 | earthquake | felt | feel |
| 10 | 3978 | weibo | circle of friends | wechat |
| 11 | 3588 | anxiety | tension | recent |
| 12 | 3393 | lose weight | body weight | terrible |
| 13 | 3063 | terrible | habit | world |
| 14 | 2986 | 2020 | 2019 | end of world |
| 15 | 2919 | dogs | dog | a |
| 16 | 2883 | wisdom teeth | tooth extraction | tooth |
| 17 | 2697 | hair | hairline | bangs |
| 18 | 2635 | teacher | start of school | school |
| 19 | 2484 | a | kitty | kitten |
| 20 | 2438 | drink | alcohol | drunk |
| 21 | 2394 | marry | fear of marriage | marriage |
| 22 | 2387 | money | some | confused |
| 23 | 2383 | save | help | child |
| 24 | 2313 | eye | eyelid | right eye |
| 25 | 2247 | subway | high speed rail | train |
| 26 | 2187 | eat | dare | terrible |
| 27 | 2054 | milk tea | coffee | cup of |
| 28 | 1974 | thunder | lightning | thunder and lightning |
| 29 | 1967 | examination | gaokao | tension |
| 30 | 1928 | go to work | not realized | jobs |
| 31 | 1901 | go out | go home | dare |
| 32 | 1888 | hospital | doctors | physical examination |
| 33 | 1880 | mosquito | mosquito coils | mosquito net |
| 34 | 1772 | guangzhou city | winter | freeze to death |
| 35 | 1689 | this year | a year | chinese new year |
| 36 | 1684 | month | july | august |
| 37 | 1680 | aircraft | airport | flight |
| 38 | 1606 | soul | hindsight | demons |
| 39 | 1588 | death | sudden death | go to die |
| 40 | 1587 | next time | na na | a bit |
| 41 | 1482 | foggy | roller coaster | climb the mountain |
| 42 | 1468 | poisonous | poisoning | mushroom |
| 43 | 1446 | nowadays | hot | tomorrow |
| 44 | 1428 | heart | heartbeat | accelerate |
| 45 | 1403 | clothes | short sleeve | long johns |
| 46 | 1395 | baby | pregnancy | doctors |
| 47 | 1376 | chongqing city | fog | hotpot |
| 48 | 1363 | elevator | power failure | stairs |

| Topic | Size | Word 1 | Word 2 | Word 3 |
|---|---|---|---|---|
| 49 | 1307 | typhoon | windy | gale |
| 50 | 1292 | sisters | sister | brother |
| 51 | 1223 | faker | bilk | phone |
| 52 | 1210 | get away | evade | unable to escape |
| 53 | 1148 | love song | rapper | a song |
| 54 | 1145 | belly | gastroscopy | gastroenteritis |
| 55 | 1130 | blood pressure | exsanguinate | hospital |
| 56 | 1126 | age | age | afraid |
| 57 | 1106 | holiday | holiday | go to work |
| 58 | 1094 | weekend | monday | friday |
| 59 | 1092 | explosion | place | flammable |
| 60 | 1083 | 30 | thirty | 25 |

## C.3 Example posts of identified non-COVID-19 related topics

**Table C.3: Example posts of identified non-COVID-19 related topics**

| Topic | Topic Keyword | Content |
|---|---|---|
| 1 | Nightmare | 做梦梦到领导……被骂醒了<br>Dreaming of the leader…woke up after being scolded |
| 1 | Nightmare | 恶梦缠身怎么治！<br>How to cure nightmare |
| 2 | Insomnia | 最怕夜幕降临，多少个不眠夜<br>I am most afraid of nightfall, so many sleepless nights. |
| 2 | Insomnia | 一直熬夜会不会猝死啊<br>Will I suddenly die if I stay up frequently |
| 12 | Lose weight | 最近不敢看的两个东西。体重秤还有日历<br>Two things that I dare not look at recently, weight and calendar |
| 12 | Lose weight | 体重噌噌噌往上涨，太可怕了<br>It is so horrible that weight is going up |
| 22 | Money | 世人慌慌张张，不过图碎银几两<br>People are panicking and hurrying around, just to earn some money |
| 22 | Money | 大灾之时，才知道余钱余粮的重要性<br>Only at the time of disasters that we know the importance of money and food. |
| 24 | Eye | 近视眼求检查，有点怕<br>I need a myopia examination, a little scary |
| 24 | Eye | 用眼过度了吗，总觉得眼睛雾蒙蒙的<br>Do I use my eys too much? I always feel that my eyes are foggy |

# D FEAR PATTERNS
## D.1 Share of attentions among non-COVID-19 related posts

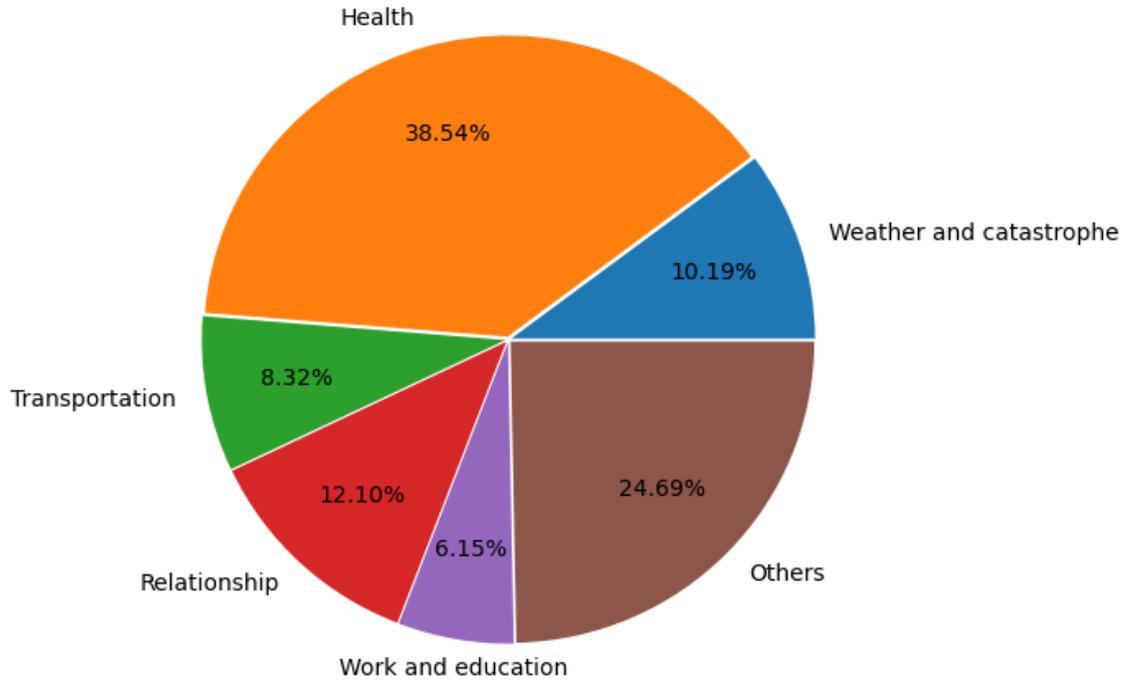

Figure D.1: Pie chart shows the attention distribution of non-COVID-19 related posts. After manually checking the most informative keywords for each cluster, we assign each clustered topic to one of the six attentions (i.e. health, weather and catastrophe, relationship, work and education, transportation and others). Among all attentions, people are more likely to be concerned with health topics.

## D.2 Intra-day temporal trend of posting nightmare and insomnia posts

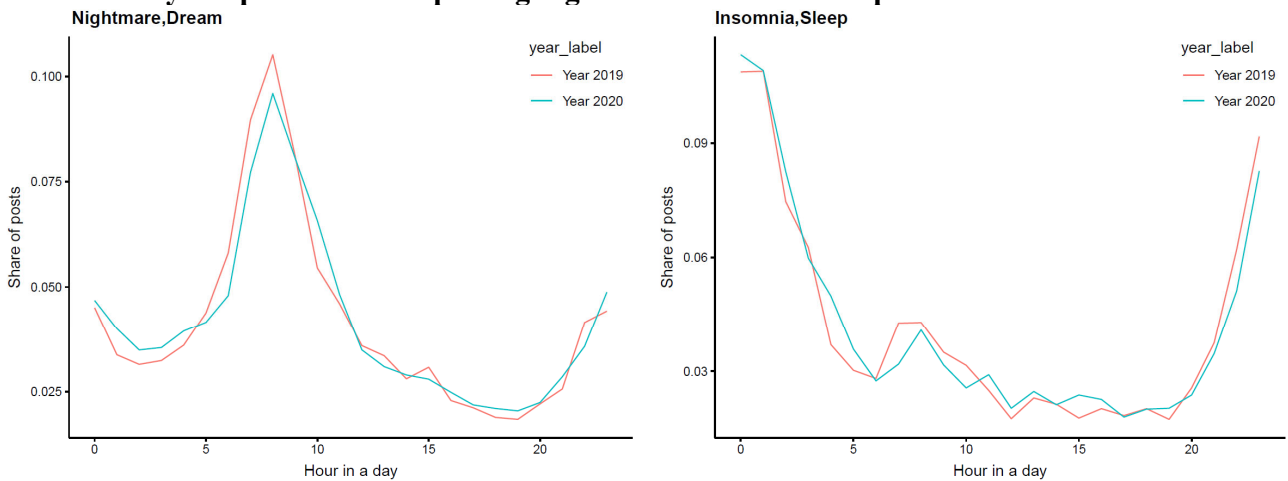

Figure D.2: Line graphs show the temporal distribution of posting a specific topic within a day by year. Neither the nightmare topic nor insomnia topic changes significantly when comparing the years 2019 and 2020, indicating the usage of the word remains constant.

## D.3 Temporal trend of average fear posts generated by different genders by week

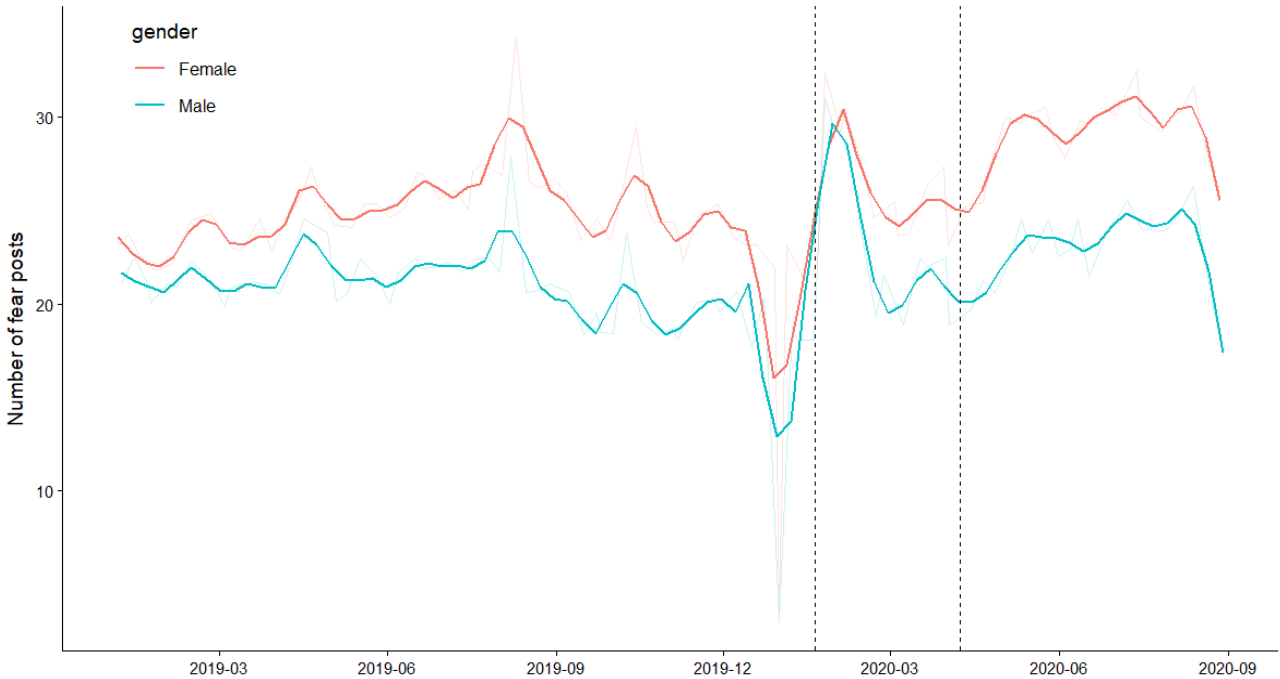

Figure D.3: Line graph shows the temporal trend of non-COVID-19 related posts by week. Each point represents the average posts number generated by 1,000 users of the same gender within that week. Female (solid line above) is more likely to post fearful contents comparing to male (solid line in the bottom) during the year 2019 and post-COVID-19 period. As shown in the graph, males reached the same level of fear posting as females at the start of COVID-19 and quickly drops.

## D.4 T-test results of topics related to health by gender

Table D.4: T-test results of topics related to health by gender

| Topic | Gender | Period | Coefficient | SE | P-Value |
|---|---|---|---|---|---|
| Nightmare, Dream | Female | COVID-19 Peak | 0.251 | 0.080 | **0.005 *** |
| Nightmare, Dream | Male | COVID-19 Peak | 0.103 | 0.077 | 0.193 |
| Nightmare, Dream | Female | Post COVID-19 | 0.270 | 0.037 | **0.000 *** |
| Nightmare, Dream | Male | Post COVID-19 | 0.197 | 0.047 | **0.000 *** |
| Insomnia, Sleep | Female | COVID-19 Peak | 0.100 | 0.050 | **0.057 *** |
| Insomnia, Sleep | Male | COVID-19 Peak | -0.016 | 0.052 | 0.759 |
| Insomnia, Sleep | Female | Post COVID-19 | 0.046 | 0.034 | 0.177 |
| Insomnia, Sleep | Male | Post COVID-19 | -0.035 | 0.036 | 0.343 |
| Cold, Fever | Female | COVID-19 Peak | -0.162 | 0.121 | 0.195 |
| Cold, Fever | Male | COVID-19 Peak | -0.047 | 0.080 | 0.568 |
| Cold, Fever | Female | Post COVID-19 | -0.118 | 0.025 | **0.000 *** |
| Cold, Fever | Male | Post COVID-19 | -0.072 | 0.023 | **0.003 *** |
| Lose weight | Female | COVID-19 Peak | -0.005 | 0.018 | 0.780 |
| Lose weight | Male | COVID-19 Peak | -0.046 | 0.021 | **0.044 ** |
| Lose weight | Female | Post COVID-19 | 0.032 | 0.018 | **0.088 *** |
| Lose weight | Male | Post COVID-19 | 0.002 | 0.015 | 0.896 |
| Eye | Female | COVID-19 Peak | -0.006 | 0.018 | 0.727 |
| Eye | Male | COVID-19 Peak | -0.035 | 0.017 | **0.045 ** |
| Eye | Female | Post COVID-19 | 0.048 | 0.011 | **0.000 *** |
| Eye | Male | Post COVID-19 | -0.009 | 0.015 | 0.575 |
| Money | Female | COVID-19 Peak | 0.034 | 0.016 | **0.051 *** |
| Money | Male | COVID-19 Peak | 0.042 | 0.022 | **0.064 *** |

| Topic | Gender | Period | Coefficient | SE | P-Value |
|---|---|---|---|---|---|
| Money | Female | Post COVID-19 | 0.062 | 0.013 | **0.000 *** |
| Money | Male | Post COVID-19 | 0.090 | 0.017 | **0.000 *** |